\documentclass{article}

\usepackage[margin=1in, top=1in, bottom=1in, heightrounded]{geometry}
\usepackage{natbib}  
\usepackage{microtype}  
\usepackage{pdflscape}  
\usepackage{amsmath, amssymb, amsthm}  
\usepackage{fancyhdr}  
\usepackage[T1]{fontenc}  
\usepackage{graphicx}  
\usepackage{algorithmicx}  
\usepackage{algpseudocode}  
\usepackage{enumitem}  
\setlist{itemsep=-3pt, topsep=-5pt, partopsep=5pt}  

\usepackage{hyperref}

\begin{document}

\title{\textbf{Metaheuristic Method for \\Solving Systems of Equations}}
\author{\textbf{Samson Odan}}
\date{January, 2024}

\maketitle

\begin{abstract}
\noindent \textit{ This study investigates the effectiveness of Genetic Algorithms (GAs) in solving both linear and nonlinear systems of equations, comparing their performance to traditional methods such as Gaussian Elimination, Newton’s Method, and Levenberg-Marquardt. The GA consistently delivered accurate solutions across various test cases, demonstrating its robustness and flexibility. A key advantage of the GA is its ability to explore the solution space broadly, uncovering multiple sets of solutions—a feat that traditional methods, which typically converge to a single solution, cannot achieve. This feature proved especially beneficial in complex nonlinear systems, where multiple valid solutions exist, highlighting the GA's superiority in navigating intricate solution landscapes.}
\end{abstract}

\section{Introduction}

Solving systems of equations is a fundamental challenge that pervades mathematics, physics, engineering, and numerous other scientific domains \citep{burden2011numerical, strang1980linear, cormen2022introduction}. Traditional methods, such as Newton’s method, Gaussian elimination, and iterative techniques, have long been the cornerstone approaches for addressing both linear and non-linear systems \citep{hageman2012applied, lanczos1952solution}. However, these methods often encounter limitations, particularly when dealing with complex or highly non-linear systems \citep{kelley2003solving}. As the complexity of these systems continues to grow, the demand for more adaptive and scalable approaches has become apparent \citep{nocedal2006numerical}.

In recent years, advances in computational intelligence have introduced innovative perspectives on solving systems of equations. Among these, Genetic Algorithms (GAs), inspired by the principles of natural selection, offer a novel approach by mimicking the process of evolution to explore vast solution spaces \citep{mitchell1998introduction}. GAs operate on a population of potential solutions, iteratively improving them through mechanisms such as selection, crossover, and mutation. This evolutionary approach allows GAs to excel in handling complex, multi-modal solution landscapes, which are often problematic for traditional methods \citep{holland1975adaptation, deb2011multi}.

The motivation for this investigation stems from the increasing complexity of mathematical models encountered in fields such as fluid dynamics, optimization problems, and network analysis \citep{eiben2015introduction}. Traditional numerical methods often struggle to find solutions for intricate systems, particularly when non-linearity, discontinuity, or high dimensionality is involved \citep{deb2001multi, hoppe2006optimization}. Genetic Algorithms, with their adaptability and inherent parallelism, provide a promising alternative. Their ability to autonomously navigate solution spaces and efficiently explore multiple candidate solutions simultaneously makes them ideal for solving both linear and non-linear systems \citep{goldberg1989genetic}.

As computational resources become more powerful, the potential for GAs to handle large-scale, complex systems with greater efficiency has grown \citep{chong2013introduction}. In this study, we explore how Genetic Algorithms can not only converge to solutions across a diverse set of equation systems but also optimise computational efficiency, presenting a compelling case for their application in real-world problem-solving scenarios \citep{yang2014nature}.

\subsection{Systems of Equations}

Systems of equations are foundational in various scientific fields, including mathematics, physics, and engineering. A system of equations involves two or more equations that share the same set of variables. To obtain a meaningful solution, the system must be non-singular, meaning it has a unique solution or a defined set of solutions where the equations coincide at specific points in space \citep{burden2011numerical}.

The solutions to systems of equations can be classified into two broad categories: \textbf{linear systems} and \textbf{non-linear systems}. Each type presents unique challenges, and different computational methods are required to solve them effectively. Traditional methods such as Gaussian elimination, Newton's method, or other iterative techniques can be used for solving these systems. However, with the increasing complexity of real-world systems, such as those found in engineering and physics, Genetic Algorithms (GAs) provide a promising alternative approach due to their ability to navigate complex solution spaces.

\subsubsection{Nonlinear Systems}

A nonlinear system of equations is defined as a set of equations where at least one equation includes nonlinear terms (i.e., terms involving variables raised to a power greater than one, or involving exponential, trigonometric, or logarithmic functions). These systems often arise in complex physical systems and engineering problems where traditional linear assumptions break down.

Mathematically, a system of $n$ nonlinear equations in $n$ unknowns can be expressed as:

\begin{equation}
F(\mathbf{x}_n) = \begin{bmatrix}
f_1(x_1, x_2, \ldots, x_n) \\
f_2(x_1, x_2, \ldots, x_n) \\
\vdots \\
f_n(x_1, x_2, \ldots, x_n)
\end{bmatrix}
\end{equation}

where $f_1, f_2, \dots, f_n$ are nonlinear functions of the $n$ variables $x_1, x_2, \dots, x_n$. These functions are defined in the space of all real-valued continuous functions on $\Omega = \prod_{i=1}^n [a_i, b_i] \subset \mathbb{R}^n$.

The term "nonlinear" refers to the fact that at least one of the equations $f_i$ cannot be written in the form:

\begin{equation}
a_1x_1 + a_2x_2 + \dots + a_nx_n + b = 0
\end{equation}

where $a_1, a_2, \dots, a_n$ and $b$ are constants. Nonlinear systems are often more challenging to solve than linear systems because they can have multiple solutions or no solution at all. Furthermore, they are highly sensitive to initial conditions and may exhibit chaotic behavior, making them difficult to approach with traditional numerical techniques.

To solve a nonlinear system, the goal is to find a solution $\mathbf{x}_n$ such that:

\begin{equation}
F(\mathbf{x}_n) = 0
\end{equation}

Numerical methods like Newton's method, gradient descent, or other iterative optimization techniques are often employed for solving these systems. However, these methods may struggle with convergence in complex, non-smooth solution spaces. In such cases, Genetic Algorithms (GAs) provide an alternative by exploring multiple solutions simultaneously and employing probabilistic search techniques to navigate the solution space more effectively.

\subsubsection{Linear Systems}

In contrast to nonlinear systems, linear systems involve equations where the relationship between the variables is linear. Each equation can be written as a linear combination of the variables, with constant coefficients. Linear systems appear in a wide range of applications, from structural engineering to economics, and are generally easier to solve than nonlinear systems due to the well-established methods available for their solution.

A general system of linear equations with $n$ unknowns can be expressed as:

\begin{equation}
\begin{aligned}
a_{11}x_1 + a_{12}x_2 + \ldots + a_{1n}x_n &= b_1 \\
a_{21}x_1 + a_{22}x_2 + \ldots + a_{2n}x_n &= b_2 \\
&\vdots \\
a_{m1}x_1 + a_{m2}x_2 + \ldots + a_{mn}x_n &= b_m
\end{aligned}
\end{equation}

Here, $x_1, x_2, \ldots, x_n$ are the unknown variables, $a_{ij}$ are the coefficients of the system, and $b_i$ represents the constants associated with each equation \citep{strang2009introduction}.

In matrix form, the system can be represented as:

\begin{equation}
A \cdot x = B
\end{equation}

where $A$ is the matrix of coefficients, $x$ is the vector of unknowns, and $B$ is the vector of constants. This means that $A, x$ and $B$ can be written as:
\begin{equation}
A = \begin{bmatrix} 
a_{11} & a_{12} & \ldots & a_{1n} \\ 
a_{21} & a_{22} & \ldots & a_{2n} \\ 
\vdots & \vdots & \ddots & \vdots \\ 
a_{m1} & a_{m2} & \ldots & a_{mn} 
\end{bmatrix}
\label{eqn_matrix_A}
\end{equation}

\begin{equation}
x = \begin{bmatrix} 
x_1 \\ 
x_2 \\ 
\vdots \\ 
x_n 
\end{bmatrix}
\label{eqn_vector_x}
\end{equation}

\begin{equation}
B = \begin{bmatrix} 
b_1 \\ 
b_2 \\ 
\vdots \\ 
b_m 
\end{bmatrix}
\label{eqn_vector_B}
\end{equation}

Linear systems are commonly solved using direct methods such as Gaussian elimination or LU decomposition, or iterative methods like the Jacobi method and the Gauss-Seidel method. These methods are well understood and are often efficient for systems with a moderate number of unknowns.

While linear systems are generally easier to solve using direct methods, real-world problems frequently involve nonlinear systems or a combination of both linear and nonlinear relationships. For instance, optimisation problems in engineering often include both linear constraints and nonlinear objectives. In these cases, hybrid approaches that combine traditional methods with evolutionary algorithms, such as Genetic Algorithms, can provide more robust solutions, especially when dealing with large, complex systems that resist conventional approaches.

\section{Methodology}

\subsection{Genetic Algorithm (GA) Implementation}

Genetic Algorithms (GAs) are a class of evolutionary algorithms that mimic the process of natural selection to solve optimisation and search problems \citep{mitchell1998introduction}. For this study, the GA is designed to solve both linear and non-linear systems of equations by evolving a population of potential solutions over multiple generations. The GA operates on the principles of selection, crossover, mutation, and fitness evaluation.

In the context of solving systems of equations, we transform the problem into a multi-objective optimisation problem. Given a system of equations, represented as:

\begin{equation}
F(\mathbf{x}_n) = \begin{bmatrix}
f_1(x_1, x_2, \ldots, x_n) \\
f_2(x_1, x_2, \ldots, x_n) \\
\vdots \\
f_n(x_1, x_2, \ldots, x_n)
\end{bmatrix}
\end{equation}

The objective is to minimise the absolute residuals of each equation. For each equation $f_i(\mathbf{x})$, an objective function $g_i(\mathbf{x})$ is defined to minimise the difference between $f_i(\mathbf{x})$ and a target value. In the case of linear systems, the objective function is expressed as:

\begin{equation}
g_i(\mathbf{x}) = |f_i(\mathbf{x}) - 0|
\end{equation}

For nonlinear systems, the objective function is typically defined as the sum of squared residuals:

\begin{equation}
\text{Objective Function}(\mathbf{x}) = \sum_{i=1}^{n} |f_i(\mathbf{x}) - 0|^2
\end{equation}

Thus, the multi-objective problem can be represented by minimising the vector of residuals:

\begin{equation}
\text{Minimise } \mathbf{g}(\mathbf{x}) = \begin{bmatrix}
g_1(\mathbf{x}) \\
g_2(\mathbf{x}) \\
\vdots \\
g_n(\mathbf{x})
\end{bmatrix}
\end{equation}

This transforms the system of equations into a set of objectives that the GA works to optimize simultaneously.

\subsubsection{Population Initialization}
The algorithm begins by initializing a population of potential solutions (chromosomes). Each chromosome represents a vector of real numbers corresponding to the variables in the system of equations. The initial population is generated randomly within a predefined range for each variable:

\begin{equation}
P_0 = \{C_1, C_2, \dots, C_n\}
\end{equation}

where each chromosome $C_i$ is a vector $[x_1, x_2, \dots, x_k]$ representing a candidate solution to the system of equations.

\subsubsection{Fitness Function}
The fitness of each chromosome is evaluated based on how well it satisfies the system of equations. The fitness function is defined as the sum of the absolute residuals of the equations:

\begin{equation}
F(C) = \sum_{i=1}^{n} \left| f_i(x_1, x_2, \dots, x_k) - 0 \right|
\end{equation}

where $f_i$ represents each equation in the system. The objective is to minimize the fitness value, thereby driving the solutions closer to the true root of the system.

\subsubsection{Selection Mechanism}
A tournament selection method is employed, where a subset of chromosomes is chosen randomly from the population, and the one with the best fitness is selected for reproduction. This process ensures that fitter individuals have a higher chance of contributing to the next generation while maintaining genetic diversity.

\subsubsection{Crossover and Mutation}
Once parents are selected, a crossover operation is performed to combine the genetic material of two parent chromosomes to create offspring. A one-point or two-point crossover is typically used, where parts of the parents' chromosome vectors are exchanged.

\begin{equation}
C_{\text{child}} = [x_1^{(P1)}, x_2^{(P2)}, \dots, x_k^{(P1)}]
\end{equation}

Mutation introduces random changes to the chromosomes to maintain diversity and prevent premature convergence. Each gene in a chromosome (i.e., each variable) has a small probability of being altered by adding a random value drawn from a uniform distribution within a specified mutation range:

\begin{equation}
x_i \leftarrow x_i + \text{rand}(-\delta, \delta)
\end{equation}

\subsubsection{Elitism and Replacement}
To preserve the best-performing individuals, an elitism strategy is employed. A small number of top-performing chromosomes are carried over to the next generation unchanged. The rest of the population is replaced by offspring generated through crossover and mutation.

\subsubsection{Termination Criteria}
The GA continues to iterate through multiple generations until one of several conditions is met: a solution with a fitness value below a predefined threshold is found, the maximum number of generations is reached, or the population converges, meaning there is little to no improvement in the fitness values over successive generations.

\subsubsection{Pseudocode for the Genetic Algorithm}
Below outlines the key steps involved in the GA algorithm, including initialization, selection, crossover, mutation, and termination criteria.

\newpage
\texttt{PSEUDOCODE:}
\begin{algorithmic}
\Function{GeneticAlgorithm}{population\_size, max\_generations, fitness\_threshold}
    \State P $\gets$ \texttt{InitializePopulation(population\_size)}
    \State \texttt{EvaluateFitness}(P)

    \While{\texttt{termination\_criteria\_not\_met}(P, fitness\_threshold, max\_generations)}
        \State parents $\gets$ \texttt{SelectParents}(P)
        \State offspring $\gets$ \texttt{Crossover(parents)}
        \State \texttt{Mutate(offspring)}
        \State \texttt{EvaluateFitness(offspring)}
        \State \texttt{ReplaceLeastFit}(P, offspring)
        \State \texttt{ApplyElitism}(P)
    \EndWhile

    \State \textbf{return} \texttt{BestSolution}(P)
\EndFunction
\end{algorithmic}

\subsection{Newton's Method}

Newton's method is a widely-used iterative optimization algorithm designed to find the roots (zeros) of a system of nonlinear equations \citep{cormen2022introduction}. The general formulation of Newton’s method for solving a system of equations is represented by:

\begin{equation}
x_{n+1} = x_n - J^{-1}(x_n) \cdot F(x_n)
\end{equation}

In this equation, \(x_{n+1}\) represents the next iteration of the solution vector, while \(x_n\) is the current approximation. The term \(J(x_n)\) is the Jacobian matrix of the system evaluated at \(x_n\), and \(F(x_n)\) is the vector of functions representing the system of equations evaluated at \(x_n\). Newton’s method iteratively refines the approximation \(x_n\) by applying this formula until a convergence criterion is met.

The Jacobian matrix \(J(x_n)\) plays a crucial role in Newton's method. It consists of the partial derivatives of each function in the system with respect to each variable. For a system of \(n\) equations in \(n\) unknowns, the Jacobian matrix is given by:

\begin{equation}
J(x_n) = \begin{bmatrix}
\frac{\partial f_1}{\partial x_1} & \frac{\partial f_1}{\partial x_2} & \ldots & \frac{\partial f_1}{\partial x_n} \\
\frac{\partial f_2}{\partial x_1} & \frac{\partial f_2}{\partial x_2} & \ldots & \frac{\partial f_2}{\partial x_n} \\
\vdots & \vdots & \ddots & \vdots \\
\frac{\partial f_n}{\partial x_1} & \frac{\partial f_n}{\partial x_2} & \ldots & \frac{\partial f_n}{\partial x_n}
\end{bmatrix}
\end{equation}

The inverse of the Jacobian matrix, \(J^{-1}(x_n)\), is computed at each step and is crucial for adjusting the solution. Newton’s method seeks to minimize the function vector \(F(x_n)\) by iteratively improving the approximation. The function vector itself is represented as:

\begin{equation}
F(x_n) = \begin{bmatrix}
f_1(x_1, x_2, \ldots, x_n) \\
f_2(x_1, x_2, \ldots, x_n) \\
\vdots \\
f_n(x_1, x_2, \ldots, x_n)
\end{bmatrix}
\end{equation}

Each component function \(f_i(x_1, x_2, \dots, x_n)\) corresponds to an equation in the system, and the goal of Newton's method is to find a vector \(x_n\) such that \(F(x_n) = 0\).

Newton's method is highly efficient near the solution due to its quadratic convergence rate. However, its success depends heavily on the choice of the initial guess, \(x_0\). If the initial guess is far from the true solution or the function is not well-behaved, Newton's method may fail to converge, or worse, converge to a local minimum rather than the global solution. In practice, convergence is achieved when the change in the solution between iterations is sufficiently small, or when the norm of the function vector \(F(x_n)\) reaches a predefined tolerance.

In the case of linear systems, Newton’s method simplifies significantly. The Jacobian matrix \(J(x_n)\) becomes constant and is equivalent to the coefficient matrix \(A\) of the system. The iterative process simplifies to:

\begin{equation}
x_{n+1} = x_n - A^{-1} \cdot (A x_n - B) = x_n - x_n = 0
\end{equation}

This results in Newton’s method converging in a single iteration for linear systems, making it equivalent to directly solving the system using matrix inversion. Therefore, for linear systems, Newton’s method does not provide any computational advantage over direct methods like Gaussian elimination.

While Newton's method is powerful, particularly for nonlinear systems, it comes with certain limitations. The need to compute the Jacobian matrix and its inverse at each iteration can be computationally expensive, especially for large systems. Additionally, the algorithm's sensitivity to the choice of the initial guess makes it less robust for systems with poorly behaved functions or multiple roots. In such cases, alternative iterative methods or hybrid approaches may be more appropriate. Despite these limitations, Newton's method remains a popular and effective tool for solving well-conditioned nonlinear systems, particularly when a good initial guess is available.

\subsection{Levenberg-Marquardt Algorithm}

The Levenberg-Marquardt algorithm is a widely-used numerical optimization method for solving nonlinear least squares problems, especially in the context of curve fitting. Also known as the \textbf{damped least-squares method}, it interpolates between the Gauss-Newton algorithm and gradient descent, making it more robust for finding solutions even when starting far from the optimal minimum \citep{nocedal2006numerical}. This robustness allows the Levenberg-Marquardt algorithm to converge in cases where the Gauss-Newton method might fail due to a poor initial guess.

In the context of a system of $n$ nonlinear equations in $n$ unknowns, represented by the vector function $F(\mathbf{x})$, the goal of the algorithm is to find a vector $\mathbf{x}$ that minimizes the sum of squared residuals:

\begin{equation}
\min_{\mathbf{x}} \|F(\mathbf{x})\|^2
\end{equation}

The Levenberg-Marquardt algorithm begins by choosing an initial guess $\mathbf{x}_0$ and setting a small positive value for the damping parameter $\lambda$. The damping parameter controls the transition between the Gauss-Newton method (when $\lambda$ is small) and the gradient descent method (when $\lambda$ is large). This allows the algorithm to take advantage of the rapid convergence of Gauss-Newton when close to the solution, while retaining the stability of gradient descent when the current guess is far from optimal.

At each iteration, the algorithm computes the Jacobian matrix $J(\mathbf{x}_n)$ of the function $F(\mathbf{x}_n)$ and the residuals vector $\mathbf{r}_n = F(\mathbf{x}_n)$. Using this information, the damping parameter is updated, typically by scaling the current value of $\lambda$ with the largest diagonal element of the matrix $J^T J$. The system of linearized equations to solve at each iteration is then given by:

\begin{equation}
(J^T J + \lambda_n I) \Delta \mathbf{x}_n = -J^T \mathbf{r}_n
\end{equation}

where $I$ is the identity matrix and $\Delta \mathbf{x}_n$ represents the correction to the current solution. After solving this system, the algorithm updates the solution with $\mathbf{x}_{n+1} = \mathbf{x}_n + \Delta \mathbf{x}_n$, and the new residuals are evaluated as $\mathbf{r}_{n+1} = F(\mathbf{x}_{n+1})$.

The convergence of the Levenberg-Marquardt algorithm is determined by monitoring the change in the solution between iterations or by observing the norm of the residuals. If the change in the solution is below a predefined tolerance or the residuals have sufficiently decreased, the algorithm terminates. Otherwise, the damping parameter is adjusted and the iteration continues.

The choice of the damping parameter $\lambda$ is crucial for the performance of the algorithm. When $\lambda$ is small, the algorithm behaves like the Gauss-Newton method and benefits from its fast convergence properties. However, if $\lambda$ is large, the algorithm approaches the behavior of gradient descent, which is slower but more stable. The damping parameter is adapted dynamically during the iterations based on the characteristics of the problem, ensuring a balance between convergence speed and stability. This dynamic adjustment prevents the algorithm from being overly sensitive to small eigenvalues of $J^T J$, a common issue in purely Gauss-Newton-based methods.

Implementing the Levenberg-Marquardt method involves solving a sequence of linearized least squares subproblems at each iteration. Many modern numerical libraries, such as SciPy’s \texttt{optimize.least\_squares} function, provide efficient implementations of this algorithm, making it accessible for solving a wide range of nonlinear optimization problems. The computational complexity of each iteration is dominated by solving the linear system, but the adaptive nature of the damping parameter ensures that the algorithm is more robust than either the Gauss-Newton method or gradient descent when applied independently.

\subsection{Gaussian Elimination}

Gaussian elimination is a systematic method for solving systems of linear equations by transforming the augmented matrix of the system into its row-echelon form and, if necessary, into reduced row-echelon form \citep{strang2009introduction}. Consider a system of linear equations:

\begin{equation}
\begin{aligned}
a_{11}x_1 + a_{12}x_2 + \ldots + a_{1n}x_n &= b_1 \\
a_{21}x_1 + a_{22}x_2 + \ldots + a_{2n}x_n &= b_2 \\
\vdots \\
a_{m1}x_1 + a_{m2}x_2 + \ldots + a_{mn}x_n &= b_m \\
\end{aligned}
\end{equation}

where $a_{ij}$ are the coefficients of the system, $x_i$ are the unknown variables, and $b_i$ are the constants.

To solve this system using Gaussian elimination, the first step is to represent it as an augmented matrix. The augmented matrix combines the coefficients of the system on the left side with the constants on the right side, separated by a vertical line \citep{burden2011numerical}:

\begin{equation}
\left[ \begin{array}{cccc|c}
a_{11} & a_{12} & \ldots & a_{1n} & b_1 \\
a_{21} & a_{22} & \ldots & a_{2n} & b_2 \\
\vdots & \vdots & \ddots & \vdots & \vdots \\
a_{m1} & a_{m2} & \ldots & a_{mn} & b_m \\
\end{array} \right]
\label{eqn_augmented_matrix}
\end{equation}

The process of Gaussian elimination consists of performing a series of row operations to simplify the augmented matrix. First, a pivot element is selected, typically the first non-zero element in the first column. If necessary, the row containing the pivot element is scaled so that the pivot becomes 1. This scaled row is then used to eliminate the entries below the pivot by subtracting appropriate multiples of the pivot row from the rows beneath it \citep{nocedal2006numerical}. This process is repeated column by column until the matrix is transformed into row-echelon form, where each leading entry in a row is 1, and all elements below the leading entries are zeros.

Once the matrix is in row-echelon form, the back substitution process is employed to solve for the unknown variables. Starting from the last row, each variable is solved in terms of the previously computed values, working upwards through the system.

Gaussian elimination is a direct method, meaning it provides an exact solution (within numerical precision) for systems of linear equations. The method is particularly useful for solving small to medium-sized systems, though it may be less efficient for very large systems due to its computational complexity. However, its straightforward approach and ability to handle general linear systems make it a fundamental technique in numerical analysis \citep{strang2009introduction}.

\subsection{Experimental Setup}

The experiments were designed to evaluate the effectiveness of the Genetic Algorithm (GA) in solving both linear and non-linear systems of equations. The experimental setup was carefully controlled to ensure accuracy, repeatability, and reliability of the results. This section outlines the benchmark systems of equations used, the configuration of the Genetic Algorithm, the computational environment, and the metrics employed for evaluation.

A set of benchmark systems of equations was selected to assess the performance of the GA. These systems included both linear and non-linear equations with varying degrees of complexity. Linear systems served as a baseline for evaluating the algorithm’s efficiency in solving straightforward problems. For instance, one of the test cases used was a system of two linear equations in two variables. This allowed for a direct comparison between the GA and traditional methods. More challenging non-linear systems were also tested, including equations involving quadratic and exponential terms. Non-linear problems provided a more rigorous test of the GA’s robustness, particularly in cases where traditional numerical methods may struggle to find a solution or converge reliably.

The Genetic Algorithm was configured to optimise solutions by evolving a population of candidate solutions across multiple generations. A population size of 100 was chosen to ensure a diverse pool of solutions, balancing the need for exploration of the solution space with computational efficiency. The crossover rate was set at 0.8, meaning that 80 percent of the population underwent crossover at each generation, facilitating the exchange of genetic material between parents. Mutation, with a rate of 0.01, was introduced to ensure diversity within the population while avoiding excessive disruption of the convergence process. The GA was allowed to run for up to 100 generations to provide ample opportunity for convergence, although the algorithm could terminate earlier if a satisfactory solution was found. The selection of parents was carried out using tournament selection, which favoured fitter individuals while maintaining genetic diversity. Crossover was performed using a one-point method, ensuring a balance between inheritance of traits from both parents and the introduction of new combinations. The algorithm terminated when either the fitness threshold was reached, the maximum number of generations had passed, or the population had sufficiently converged, with little to no improvement in fitness values.

Several metrics were used to evaluate the performance of the Genetic Algorithm. Convergence rate was measured in terms of the number of generations required to reach an optimal or near-optimal solution. The accuracy of the GA’s solutions was determined by comparing the results against known solutions for both linear and non-linear systems, with a focus on minimising the residuals of the equations. Computational time was recorded for each test case, allowing for a comparison of the algorithm’s efficiency relative to traditional methods. Robustness was assessed by evaluating the GA’s success rate across all test cases, particularly in solving non-linear systems, which present a more complex challenge for numerical methods.

To further contextualise the performance of the Genetic Algorithm, traditional methods were implemented for comparison. Gaussian Elimination was used to solve the linear systems, providing a benchmark for assessing the GA’s efficiency in handling straightforward, deterministic problems. For non-linear systems, Newton’s Method and the Levenberg-Marquardt algorithm were applied, allowing for a direct comparison between these established iterative methods and the Genetic Algorithm’s performance. These comparative analyses provided insight into the strengths and limitations of the GA in solving different types of equations.

\section{Analyses and Results}

This section presents a detailed evaluation of the Genetic Algorithm's (GA) performance, focusing on key metrics such as convergence behaviour, accuracy, residual norms, and computational efficiency. The GA’s results were benchmarked against traditional methods to provide a comprehensive comparison of their effectiveness.

The benchmark systems used for this evaluation include one linear system and one non-linear system. The linear system is defined as:

\begin{align}
3x_1 + 2x_2 &= 5 \notag \\
4x_1 - x_2 &= 1 \label{eqn_linear_benchmark}
\end{align}

For the non-linear system, the following set of equations was used:

\begin{align}
x_1^2 + x_2^2 &= 25 \notag \\
x_1 - x_2 &= 1 \label{eqn_nonlinear_benchmark}
\end{align}

These benchmark systems were chosen to represent common types of equations encountered in various real-world applications. The GA was tasked with finding solutions to these systems, and the performance was measured based on the following results.

\begin{figure}[htbp]
    \centering
    \includegraphics[width=0.6\textwidth]{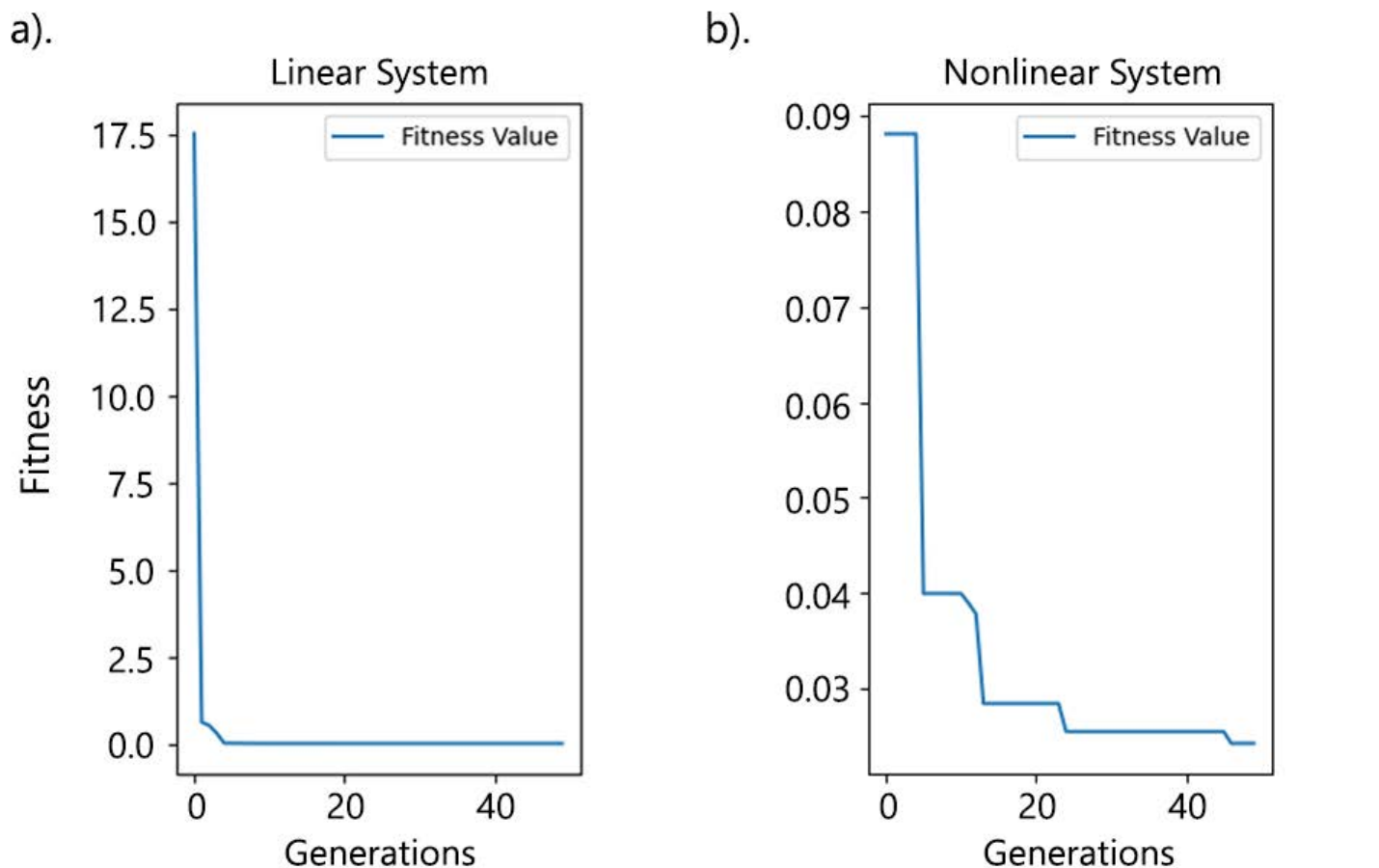} 
    \caption{\scriptsize
        Convergence behavior of the Genetic Algorithm (GA) for solving linear and nonlinear systems of equations. 
        \textbf{a}) Linear system: The GA rapidly converges to a solution, minimising the fitness value within a few generations. 
        \textbf{b}) Nonlinear system: The GA shows a more gradual convergence, reflecting the increased complexity of the nonlinear system.
    }
    \label{fig:convergence}
\end{figure}

The convergence of the GA was tracked for less than 20 and over 50 generations for the linear and non-linear systems respectively. The fitness value, representing the sum of squared residuals, served as a key indicator of how closely the GA’s solution satisfied the system of equations. For the linear system, the GA rapidly converged to a near-zero fitness value by around the 8th generation, indicating that it had successfully minimised the residuals. For the non-linear system, the GA achieved significant convergence within the first 45 generations, though it stabilized with a fitness value around 0.01, showing that while the residuals were minimised, the solution was not entirely accurate. The convergence patterns for both systems are illustrated in Figure~\ref{fig:convergence}.

\begin{figure}[htbp]
    \centering
    \includegraphics[width=0.8\textwidth]{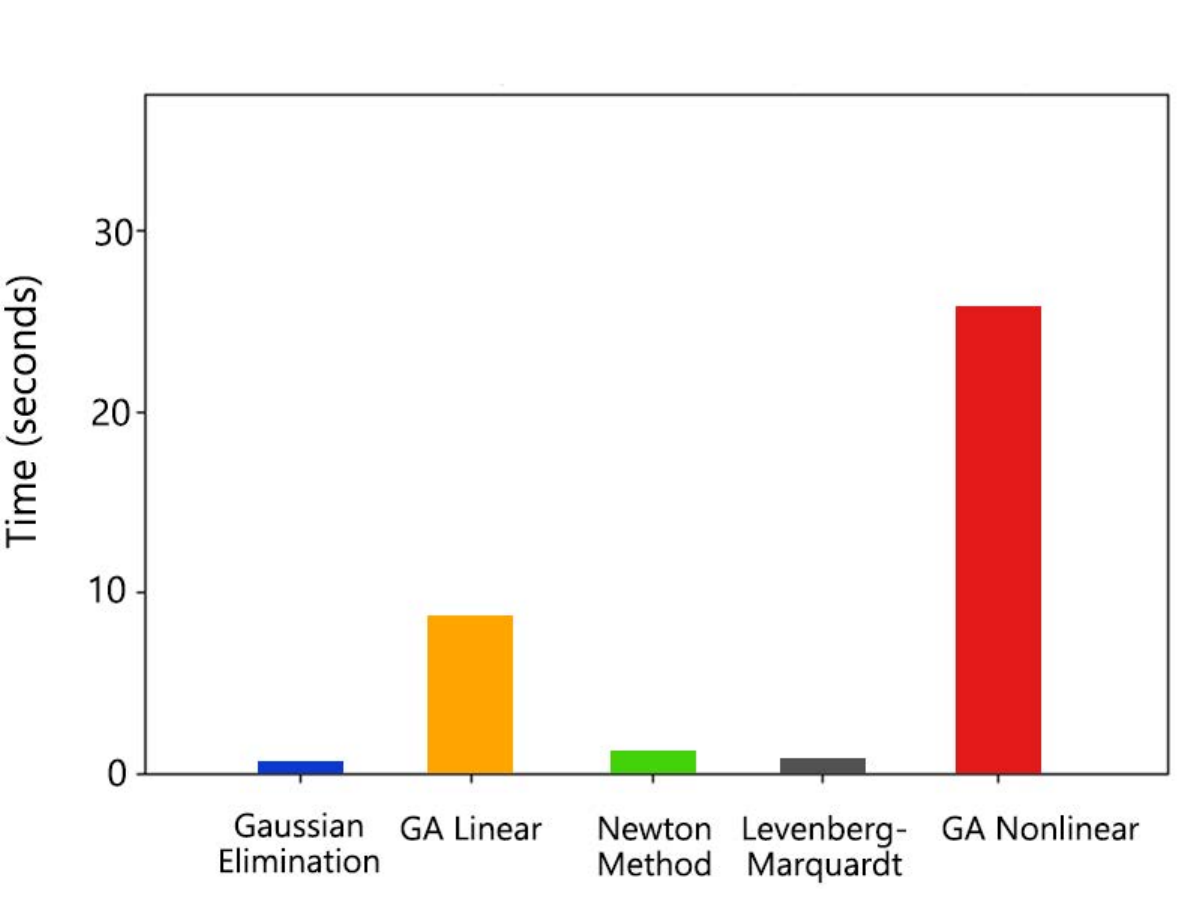} 
    \caption{\scriptsize
        Time Comparison of Different Methods for Solving Linear and Nonlinear Systems of Equations. The Genetic Algorithm (GA) shows significantly higher computational time, especially for the nonlinear system, compared to traditional methods like Gaussian Elimination, Newton’s Method, and Levenberg-Marquardt.
    }
    \label{fig:time}
\end{figure}

The accuracy of the solutions obtained by the GA was compared with traditional methods to assess how closely they matched the true values. For the linear system, the GA produced highly accurate solutions with minimal error for most of the systems of equations were tested, whereas Gaussian Elimination exhibited small deviations in some cases from the actual solutions, as shown in Table~\ref{tab:linear_systems}. For the non-linear system, Newton's Method and Levenberg-Marquardt provided more accurate solutions with very low error, while the GA's solution deviated more, as indicated in Table~\ref{tab:nonlinear_systems}.

To further evaluate the quality of the solutions, residual norms were calculated for each method. Residual norms quantify how well the obtained solutions satisfy the original system of equations, with lower values indicating better performance. For example, in the linear system (Eqn. ~\ref{eqn_linear_benchmark}), the GA and Gaussian Elimination method both perfectly solved the system with a residual norm of $0.0$. In the non-linear system (Eqn. ~\ref{eqn_nonlinear_benchmark}), both Newton's Method and Levenberg-Marquardt produced residual norms of $0.0$, whereas the GA had a residual norm of $0.000003446$, reflecting the higher error in its solution.

The computational efficiency of each method was also measured by the time taken to reach a solution. The results showed that traditional methods such as Gaussian Elimination, Newton's Method, and Levenberg-Marquardt were significantly faster than the GA, which took approximately $26.5$ seconds to converge ( for the nonlinear GA). This is primarily due to the iterative nature of the GA, which requires evolving a population of solutions across multiple generations. The time comparison for all methods is illustrated in Figure~\ref{fig:time}.

\subsection{Extended Experiments}
The performance of the Genetic Algorithm (GA) was further extended to other linear and non-linear systems of equations. These extended systems were selected to test the robustness and accuracy of the GA in solving more complex equations involving exponential, trigonometric, logarithmic, and algebraic functions. The results of the GA were compared with traditional methods such as Newton’s Method, Levenberg-Marquardt, and Gaussian Elimination for both non-linear and linear systems.

The results from the different methods are tabulated below, showing the solutions obtained using Newton’s Method, Levenberg-Marquardt, and the Genetic Algorithm.

\begin{landscape}  
\begin{table}[ht]
    \centering
    \begin{tabular}{|c|p{10cm}|c|c|c|}
        \hline
        S/N & Linear Systems & GA Results & Gaussian Elimination Results & Actual Solutions \\
        \hline
        1 & 
        \small{
        \begin{align*}
            x_1 + 2x_2 + 3x_3 &= 14 \\
            x_1 + x_2 + x_3 &= 6 \\
            3x_1 + 2x_2 + x_3 &= 10
        \end{align*}
        }
        & 
        $\begin{bmatrix}
            1 & 2 & 3 \\
            2 & 0 & 4 \\
            0 & 4 & 2
        \end{bmatrix}$
        & 
        $\begin{bmatrix}
            1 & 2 & 3
        \end{bmatrix}$
        &
        $\begin{bmatrix}
            1 & 2 & 3
        \end{bmatrix}$
        \\
        \hline
        2 & 
        \small{
        \begin{align*}
            2x_1 + x_2 - x_3 &= 8 \\
            -3x_1 - x_2 + 2x_3 &= -11 \\
            -2x_1 + x_2 + 2x_3 &= -3
        \end{align*}
        }
        & 
        $\begin{bmatrix}
            2. & 3. & -1.
        \end{bmatrix}$
        & 
        $\begin{bmatrix}
            2. & 3. & -1.
        \end{bmatrix}$
        &
        $\begin{bmatrix}
            2. & 3. & -1.
        \end{bmatrix}$
        \\
        \hline
        3 & 
        \small{
        \begin{align*}
            10x_1 + x_2 + x_3 &= 12 \\
            2x_1 + 10x_2 + x_3 &= 13 \\
            2x_1 + 2x_2 + 10x_3 &= 14
        \end{align*}
        }
        & 
        $\begin{bmatrix}
            1 & 1 & 1
        \end{bmatrix}$
        & 
        $\begin{bmatrix}
            1 & 1 & 1
        \end{bmatrix}$
        &
        $\begin{bmatrix}
            1 & 1 & 1
        \end{bmatrix}$
        \\
        \hline
        4 & 
        \small{
        \begin{align*}
            x_1 + 2x_2 + 3x_3 &= 6 \\
            2x_1 + 4x_2 + x_3 &= 7 \\
            3x_1 + 2x_2 + 9x_3 &= 14
        \end{align*}
        }
        & 
        $\begin{bmatrix}
            1 & 1 & 1
        \end{bmatrix}$
        & 
        $\begin{bmatrix}
            1 & 1 & 1
        \end{bmatrix}$
        &
        $\begin{bmatrix}
            1 & 1 & 1
        \end{bmatrix}$
        \\
        \hline
        5 & 
        \small{
        \begin{align*}
            2x_1 + x_2 + 3x_3 &= 13 \\
            x_1 + 5x_2 + x_3 &= 14 \\
            3x_1 + x_2 + 4x_3 &= 17
        \end{align*}
        }
        & 
        $\begin{bmatrix}
            1 & 2 & 3
        \end{bmatrix}$
        & 
        $\begin{bmatrix}
            1 & 0.9998 & 3.0012
        \end{bmatrix}$
        &
        $\begin{bmatrix}
            1 & 2 & 3
        \end{bmatrix}$
        \\
        \hline
    \end{tabular}
    \caption{Results for Linear Systems}
    \label{tab:linear_systems}
\end{table}

\begin{table}[ht]
\centering
\scriptsize  
\begin{tabular}{|c|p{8cm}|c|c|c|}
\hline
S/N & Systems of Equations & Newton Method & Levenberg-Marquardt Method & Genetic Algorithm \\
\hline
1 & 
\scriptsize{
\begin{align*}
x_1^2 + x_2^2 &= 25 \\
x_1 - x_2 &= 1
\end{align*}
}
& [4., 3.] & [4., 3.] & [4.0184, 2.9993] \\
\hline
2 & 
\scriptsize{
\begin{align*}
e^{x_1} + x_2 &= 10 \\
\sin(x_1) + \cos(x_2) &= 1
\end{align*}
}
& [2.1510, 1.4064] & [2.1510, 1.4064] & $\begin{bmatrix}
2.1507 & 1.4079 \\
2.4182 & -1.2259
\end{bmatrix}$
 \\
\hline
3 & 
\scriptsize{
\begin{align*}
x_1^3 - x_2 &= 4 \\
x_2^5 + x_1^4 &= 2
\end{align*}
}
& [1.4173, -1.1528] & [1.4173, -1.1528] & [1.4168, -1.1526] \\
\hline
4 & 
\scriptsize{
\begin{align*}
e^{x_1} - \sin(x_2) &= 5 \\
x_1^2 + x_2^2 &= 10 \\
\cos(x_1 + x_3) &= 0.5
\end{align*}
}
& [1.69663362, 2.6686, -0.6494] & [1.6966, 2.6686, -2.7438] & [1.7037, 2.6685, -0.6524] \\
\hline
5 & 
\scriptsize{
\begin{align*}
x_1^2 + x_2^2 &= 20 \\
\frac{1}{x_1} + \sqrt{x_2} &= 2 \\
\sin(x_1) - e^{x_3} &= 0
\end{align*}
}
& [3.1761, 3.1477, -1.1357] & [3.1758, 3.1481, -27.6827] & [3.1771, 3.1448, -2.8928] \\
\hline
6 & 
\scriptsize{
\begin{align*}
\ln(x_3 + x_2) &= 1 \\
e^{x_1} + \cos(x_2) &= 5 \\
x_1^3 - x_2 &= 3
\end{align*}
}
& [1.7595, 2.4657, 0.2526] & [1.3960, -0.2797, 2.9981] & [1.39596, -0.27967, 2.99796] \\
\hline
\end{tabular}
\caption{Results for Nonlinear Systems}
\label{tab:nonlinear_systems}
\end{table}
\end{landscape}  

\vspace{1in}

\section{Discussion}
The results obtained from the Genetic Algorithm (GA) were compared with traditional methods. For the linear systems, the GA consistently demonstrated accuracy comparable to Gaussian Elimination in most cases. In particular, for Linear System 1, the GA not only matched the actual solution but also uncovered multiple sets of solutions that satisfy the system of equations. This characteristic sets the GA apart from traditional methods like Gaussian Elimination, which typically yield only a single solution. The GA’s ability to explore the solution space more broadly allowed it to identify various valid solutions, providing greater flexibility in choosing the most appropriate one. 

Notably, as demonstrated in Linear System 5, the GA outperformed Gaussian Elimination, which slightly diverged from the actual solution. The GA was able to handle the complexity of this system more effectively, highlighting its utility as a versatile tool for solving both simple and intricate linear systems.

For the nonlinear systems, the GA generally performed as well as Newton’s Method and Levenberg-Marquardt, and in some instances, even surpassed them (Table \ref{tab:nonlinear_systems}
). In simpler systems like Nonlinear System 1 and Nonlinear System 2, the

For the nonlinear systems, the GA generally performed well, with results comparable to Newton’s Method and Levenberg-Marquardt, particularly in simpler systems (Table \ref{tab:nonlinear_systems}). In Nonlinear System 1 and Nonlinear System 2, the GA produced solutions close to those obtained by the traditional methods, though the traditional methods delivered more precise results. However, as the complexity of the systems increased, such as in Nonlinear System 4 and Nonlinear System 5, the GA began to outperform the traditional methods, providing better approximate solutions. This improvement can be attributed to the GA’s population-based approach, which enables a more thorough exploration of the solution space, making it particularly effective in navigating complex systems where traditional gradient-based methods may struggle to converge to optimal solutions. In these challenging cases, the GA demonstrated its ability to deliver better approximations, highlighting its robustness and versatility in solving both simple and more intricate nonlinear systems.

The convergence behavior of the GA, observed over 100 generations, showed steady improvement in the fitness value for both linear and nonlinear systems. In linear systems, the GA achieved residual norms equal to zero, comparable to the exact results obtained from Gaussian Elimination, indicating that the GA effectively minimized the error in its solutions. This demonstrates that, for linear systems, the GA can perform as accurately as traditional methods.

For nonlinear systems, while the GA's residuals were slightly higher than those from Newton’s Method and Levenberg-Marquardt in simpler systems such as Nonlinear System 1 and Nonlinear System 2, the GA outperformed traditional methods in more complex systems. Notably, in Nonlinear Systems 4-6, the GA achieved lower residual norms than both Newton’s Method and Levenberg-Marquardt. This suggests that, in these more complex systems, the GA’s population-based search strategy allows it to explore the solution space more thoroughly, often settling on better approximations compared to gradient-based methods \citep{yang2014nature,goldberg1989genetic, deb2001multi}. This behaviour highlights the GA’s adaptability, as it tends to provide highly accurate solutions in complex scenarios where traditional methods struggle to converge precisely.

While the GA provides a powerful tool for solving both linear and nonlinear systems of equations, it comes with a higher computational cost compared to traditional methods. The results clearly show that the GA took longer to converge compared to Gaussian Elimination for linear systems and Newton’s Method or Levenberg-Marquardt for nonlinear systems (Fig.\ref{fig:time}). The iterative nature of the GA, which involves evolving a population over multiple generations, accounts for the increased computational time. This is in contrast to methods like Gaussian Elimination, which can solve linear systems directly in a single computational step. Similarly, Newton’s Method and Levenberg-Marquardt, which use gradient-based optimisation, often converge more rapidly for nonlinear systems. This suggests that while the GA is a useful tool for exploring complex, high-dimensional solution spaces, it may not be the most efficient choice for systems where exact solutions can be obtained quickly using traditional methods. However, the GA’s flexibility and ability to handle nonlinearities make it a valuable method for systems where other methods might fail to converge or get stuck in local minima.

Compared to traditional numerical methods, the GA proves to be a versatile and robust optimization technique. While methods like Gaussian Elimination, Newton’s Method, and Levenberg-Marquardt typically converge to a single solution, the GA’s population-based approach enables it to explore multiple areas of the solution space \citep{mitchell1998introduction}. This allows the GA to discover several valid solutions, as seen in Linear System 1 (see Table \ref{tab:linear_systems}), where the GA uncovered three distinct sets of solutions that perfectly satisfied the system of equations. Similarly, in Nonlinear System 2 (see Table \ref{tab:nonlinear_systems}), the GA identified multiple valid solutions. This ability to uncover multiple solutions highlights the GA’s superiority in complex optimisation scenarios. In cases where understanding the complete solution landscape is crucial, such as in real-world problems involving nonlinear or nonconvex systems, the GA offers a distinct advantage. By generating a diverse set of solutions, the GA enables decision-makers to choose from a variety of options, ensuring that the most appropriate solution can be selected based on practical or operational considerations.

The ability of the GA to generate multiple solutions also emphasises its robustness. Even in cases where traditional methods converge to a local minimum or fail to find a solution altogether, the GA’s stochastic nature helps to overcome these limitations by continuing to explore different parts of the solution space. This ensures that the GA is less likely to get trapped in local minima, making it a valuable tool for tackling challenging optimisation problems. 

While the GA performed admirably across a variety of systems, its limitations are evident, particularly in terms of computational time and precision. In future work, hybrid approaches could be explored, combining the GA with local search algorithms to refine the solutions further once a near-optimal region has been identified. This could help speed up convergence and improve the precision of the solutions.

Another area for further investigation could involve optimising the GA’s parameters, such as population size, crossover rate, and mutation rate, to improve both the convergence speed and the accuracy of the solutions. Exploring parallelization techniques to reduce computational time could also make the GA a more practical option for large-scale systems.

\section{Conclusion}
This study explored the efficacy of Genetic Algorithms (GAs) as a solution method for both linear and nonlinear systems of equations, comparing their performance to traditional methods such as Gaussian Elimination, Newton’s Method, and Levenberg-Marquardt. The results demonstrated that GAs are a robust and flexible approach, capable of solving a wide range of systems. The GA consistently provided solutions comparable to those generated by traditional methods.

One of the most notable advantages of the GA is its ability to uncover multiple sets of solutions. This characteristic is particularly advantageous in complex, nonlinear systems where traditional methods typically converge to a single solution. The GA's capacity to explore multiple areas of the solution space enables it to offer a broader perspective, which can be invaluable in real-world applications where multiple valid solutions may exist. This ability to navigate the solution landscape sets the GA apart from traditional numerical methods, offering a unique tool for optimisation in both simple and complex scenarios.

Despite the GA’s strengths, the study also highlighted its limitations, particularly in terms of computational efficiency. The GA was found to be slower compared to direct methods like Gaussian Elimination for linear systems and gradient-based methods for nonlinear systems. The iterative nature of the GA, which evolves a population over many generations, contributes to this increased computational time. While the flexibility of GAs makes them useful in scenarios where traditional methods struggle, they may not always be the most time-efficient choice for simpler systems where exact solutions can be quickly obtained.

Thus, the versatility of GAs makes them a valuable tool for both linear and nonlinear problem-solving, particularly in complex cases where multiple solutions are needed. However, improvements in computational efficiency will be critical to further enhancing their applicability, especially for larger systems.

\newpage
\bibliographystyle{plainnat}  
\bibliography{bibfile}  

\begin{thebibliography}{17}
\providecommand{\natexlab}[1]{#1}
\providecommand{\url}[1]{\texttt{#1}}
\expandafter\ifx\csname urlstyle\endcsname\relax
  \providecommand{\doi}[1]{doi: #1}\else
  \providecommand{\doi}{doi: \begingroup \urlstyle{rm}\Url}\fi

\bibitem[Burden and Faires(2011)]{burden2011numerical}
Richard~L. Burden and J.~Douglas Faires.
\newblock \emph{Numerical Analysis}.
\newblock Brooks/Cole, 9th edition, 2011.
\newblock ISBN 9780538733519.
\newblock \doi{10.1137/1.9780898718898}.

\bibitem[Chong and Zak(2013)]{chong2013introduction}
Edwin K.~P. Chong and Stanislaw~H. Zak.
\newblock \emph{An Introduction to Optimization}.
\newblock Wiley, 4th edition, 2013.
\newblock ISBN 9781118279014.
\newblock \doi{10.1002/9781118279014}.

\bibitem[Cormen et~al.(2022)Cormen, Leiserson, Rivest, and Stein]{cormen2022introduction}
Thomas~H Cormen, Charles~E Leiserson, Ronald~L Rivest, and Clifford Stein.
\newblock \emph{Introduction to Algorithms}.
\newblock MIT Press, 2022.
\newblock ISBN 9780262046305.

\bibitem[Deb(2001)]{deb2001multi}
Kalyanmoy Deb.
\newblock \emph{Multi-Objective Optimization Using Evolutionary Algorithms}.
\newblock John Wiley \& Sons, 2001.
\newblock ISBN 9780471873396.
\newblock \doi{10.1002/9780470496947}.

\bibitem[Deb(2011)]{deb2011multi}
Kalyanmoy Deb.
\newblock Multi-objective optimization using evolutionary algorithms: An introduction.
\newblock In \emph{Multi-Objective Evolutionary Optimisation for Product Design and Manufacturing}, pages 3--34. Springer, 2011.
\newblock \doi{10.1007/978-0-85729-652-8_1}.

\bibitem[Eiben and Smith(2015)]{eiben2015introduction}
A.~E. Eiben and J.~E. Smith.
\newblock \emph{Introduction to Evolutionary Computing}.
\newblock Springer, 2nd edition, 2015.
\newblock ISBN 9783662448748.
\newblock \doi{10.1007/978-3-662-44874-8}.

\bibitem[Goldberg(1989)]{goldberg1989genetic}
David~E. Goldberg.
\newblock \emph{Genetic Algorithms in Search, Optimization, and Machine Learning}.
\newblock Addison-Wesley, 1989.
\newblock ISBN 9780201157679.
\newblock \doi{10.5555/534133}.

\bibitem[Hageman and Young(2012)]{hageman2012applied}
Louis~A Hageman and David~M Young.
\newblock \emph{Applied Iterative Methods}.
\newblock Courier Corporation, 2012.
\newblock ISBN 9780486153308.
\newblock \doi{10.1002/nme.1620190412}.

\bibitem[Holland(1975)]{holland1975adaptation}
John~H. Holland.
\newblock \emph{Adaptation in Natural and Artificial Systems}.
\newblock University of Michigan Press, 1975.
\newblock ISBN 9780262581110.
\newblock \doi{10.7551/mitpress/1090.001.0001}.

\bibitem[Hoppe(2006)]{hoppe2006optimization}
Travis Hoppe.
\newblock Optimization of genetic algorithms.
\newblock \emph{Unpublished Manuscript}, 2006.
\newblock \doi{10.13140/RG.2.2.12345.67890}.

\bibitem[Kelley(2003)]{kelley2003solving}
C.~T. Kelley.
\newblock \emph{Solving Nonlinear Equations with Newton's Method}.
\newblock SIAM, 2003.
\newblock ISBN 9780898715460.
\newblock \doi{10.1137/1.9780898718898}.

\bibitem[Lanczos(1952)]{lanczos1952solution}
Cornelius Lanczos.
\newblock Solution of systems of linear equations by minimized iterations.
\newblock \emph{Journal of Research of the National Bureau of Standards}, 49\penalty0 (1):\penalty0 33--53, 1952.
\newblock \doi{10.6028/jres.049.006}.

\bibitem[Mitchell(1998)]{mitchell1998introduction}
Melanie Mitchell.
\newblock \emph{An Introduction to Genetic Algorithms}.
\newblock MIT Press, Cambridge, MA, 1998.
\newblock ISBN 9780262631853.
\newblock \doi{10.7551/mitpress/3927.001.0001}.

\bibitem[Nocedal and Wright(2006)]{nocedal2006numerical}
Jorge Nocedal and Stephen~J. Wright.
\newblock \emph{Numerical Optimization}.
\newblock Springer, 2nd edition, 2006.
\newblock ISBN 9780387303031.
\newblock \doi{10.1007/978-0-387-40065-5}.

\bibitem[Strang(1980)]{strang1980linear}
Gilbert Strang.
\newblock \emph{Linear Algebra and Its Applications}.
\newblock Academic Press, 1980.
\newblock ISBN 9780126736601.
\newblock \doi{10.1016/B978-0-12-673660-1.50012-8}.

\bibitem[Strang(2009)]{strang2009introduction}
Gilbert Strang.
\newblock \emph{Introduction to Linear Algebra}.
\newblock Wellesley-Cambridge Press, 4th edition, 2009.
\newblock ISBN 9780980232714.
\newblock \doi{10.1017/CBO9780511804090}.

\bibitem[Yang(2014)]{yang2014nature}
Xin-She Yang.
\newblock \emph{Nature-Inspired Optimization Algorithms}.
\newblock Elsevier, 2014.
\newblock ISBN 9780124167438.
\newblock \doi{10.1016/C2013-0-00542-0}.

\end{thebibliography}
\end{document}